\documentclass[11pt,a4paper]{article}
\usepackage[hyperref]{emnlp-ijcnlp-2019}
\usepackage{times}
\usepackage{latexsym}

\usepackage{amsmath}
\usepackage{multirow}
\usepackage{multicol}
\usepackage{graphicx}
\usepackage{adjustbox}
\usepackage{arydshln}
\usepackage{comment}
\usepackage{caption, subcaption}

\usepackage[toc,page]{appendix}

\usepackage{enumerate}
\usepackage{color}

\DeclareMathOperator*{\argmin}{argmin}

\newcommand{\norm}[1]{\left\lVert#1\right\rVert_F}

\usepackage{url}

\aclfinalcopy 

\newcommand\confname{EMNLP-IJCNLP 2019}

\title{Instructions for \confname{} Proceedings}

\title{Balancing the composition of word embeddings\\ across heterogenous data sets}

\author{
  Stephanie Brandl,~~~David Lassner,~~~Maximilian Alber \\
  \texttt{[stephanie.brandl, lassner, maximilian.alber]@tu-berlin.de}\\
  \\
  Technische Universit\"at Berlin, 10623 Berlin, Germany\\
  Machine Learning Group\\
}

\date{}

\begin{document}
\maketitle
\begin{abstract}
Word embeddings capture semantic relationships based on contextual information and
are the basis for a wide variety of natural language processing applications. Notably these relationships are solely learned from the data
and subsequently the data composition impacts the semantic of embeddings
--- which arguably can lead to biased word vectors.
Given qualitatively different data subsets,
we aim to align the influence of single subsets on the resulting word vectors,
while retaining their quality.
In this regard we propose a criteria to measure the shift towards a single data subset
and develop approaches to meet both objectives.
We find that a weighted average of the two subset embeddings balances the influence of those subsets while word similarity performance decreases. We further propose a promising optimization approach to balance influences and quality of word embeddings.
\end{abstract}

\section{Introduction}

The advent of word embeddings~\citep{word2vec, pennington2014} has shifted the entire field of Natural Language Processing~(NLP) from sparse representations, such as Bag-of-Words, to dense, vectorial representations that have proven to be capable of capturing meaningful syntactic and semantic concepts.
Word embeddings are widely used in, e.g., text classification~\citep{joulin2016} and machine translations~\citep{mikolov2013b}.
Subsequently word embeddings have a crucial impact on downstream applications and,
moreover, such models inherit (hidden) assumptions and properties of the data.

Text corpora for training word embeddings are typically composed of subsets with different properties.
Properties can manifest, e.g., as U.K./U.S. English, 
but can also be induced by the authors, e.g., texts written by different genders, in different periods of time or in different contexts such as arts and politics.
While it is the intention in the first place to capture semantic and syntactic information from the data in the best possible way
--- ideally by learning from as much data as possible,
we argue that on second thought it is desirable to influence the composition of data (sub)sets.

Given a corpus where one category outnumbers the other, joint word embeddings will expose a bias towards the former --- yet this might not reflect (actual) word semantics appropriately or can be simply undesired.

The desideratum would be for example that in a transfer learning setting word embeddings trained on a large data set are to be fine-tuned on a small task-specific data set. Or in order to achieve semantics of cultural diversity, several smaller newspaper data sets with different foci could be added to a large base newspaper data set with a Euro/US centered focus.

The problem of bias increases for word embeddings as they are often used as a starting point in e.g.\ downstream tasks. Those methods usually work in a black box manner whose decision making is  difficult to see through. 

Typical state-of-the-art embedding learning algorithms do not distinguish between different data subsets and
thus merge their properties in an incidental manner.
A notable exception is the work of \citet{goikoetxea2016single} that shows how text-based and wordnet-based~\cite{Miller1995} embeddings can be combined to improve the embedding quality,
yet does not align the contribution of the individual data sets. For more details on related work we refer to Appendix \ref{appendix:rw}.

In this contribution we research if and how the influence of individual subsets can be aligned,
while retaining embedding quality w.r.t.\ word vectors learned on all the data.
For this aim we propose a measure for the retained semantics of a subset in the final embedding and compare a total of 9 different combination methods (1-9) which are explained in detail in Section \ref{sec:methods}. The combinations vary in that they are (1) trained on the complete data set, (2-4) created \textit{without}~\cite{goikoetxea2016single},
and (5-9) \textit{with} consideration of the data distribution (our approaches).

\section{Related work}
\label{appendix:rw}
 Various authors combined text-based word embeddings with additional resources, as for instance wordnet-based information, embeddings trained by different algorithms or additional data sets \citep{goikoetxea2016single, rothe2017autoextend, speer2016ensemble, Henriksson2014}. The main goal in those articles is to improve the quality of word embeddings overall.\\
 However, to the best of our knowledge, so far no one adressed the influence subsets have on a combined embedding systematically in order to balance the impact of different data sets after their composition, while retaining the quality of the word embeddings. 

\section{Evaluating the influence of data subsets on word embeddings}\label{sec:evaluation}

\begin{table*}[t]
\centering
\small
\begin{tabular}{| l | c | c | c | c || p{.5cm} | p{.6cm} | p{.5cm} | c | c | c | c || p{.6cm} | p{.6cm} | p{.6cm} | }
    \hline
    & \multicolumn{7}{c|}{New York Times}&  \multicolumn{7}{c|}{Wikipedia}\\
    \hline & & & &  & \multicolumn{3}{c|}{Analogy-test (in \%)}& & & & & \multicolumn{3}{c|}{Analogy-test (in \%)}\\ \cline{6-8} \cline{13-15}
    & $\mathcal{J}^{90}$ & $\mathcal{J}^{00}$ & $\mathcal{J}^\Delta$ & $\bar{\mathcal{J}}$ & \;\;n=1\;\; & \;\;n=5\;\; & \;\;n=10\;\; & $\mathcal{J}^{arts}$ & $\mathcal{J}^{pol}$ & $\mathcal{J}^\Delta$ & $\bar{\mathcal{J}}$ & \;\;n=1\;\; & \;\;n=5\;\; & \;\;n=10\;\;\\ \hline \hline

    (a) $U_{s}$       & 1.00  & 0.15  & 0.85              & \textbf{0.57}     & 1.61          & 6.82              & 9.33              & 1.00  & 0.20  & 0.80              & \textbf{0.60}     & 7.47              & 29.05             & 36.69             \\
    (b) $U_{l}$       & 0.15  & 1.00  & -0.85             & \textbf{0.57}     & 1.55          & 9.66              & 13.29             & 0.20  & 1.00  & -0.80             & \textbf{0.60}     & 19.37             & 53.54             & 62.12             \\\hdashline
    (1) $U_{s/l}$     & 0.20  & 0.42  & -0.21             & 0.31              & \textit{3.82} & \textit{16.05}    & \textit{20.91}    & 0.27  & 0.55  & -0.27             & 0.41              & \textbf{21.21}    & \textit{60.58}    & \textbf{69.13}    \\ \hline
    (2) AVG           & 0.24  & 0.44  & -0.21             & \textit{0.34}     & \textit{2.67} & 11.24             & 15.31             & 0.32  & 0.54  & -0.22             & \textit{0.43}     & \textit{19.21}    & \textit{53.15}    & \textit{62.20}    \\
    (3) CON           & 0.30  & 0.39  & -0.09             & \textit{0.34}     & 1.61          & \textit{12.16}    & \textit{16.03}    & 0.39  & 0.45  & -0.06             & 0.42              & 13.70             & 52.78             & 61.71             \\
    (4) PCA           & 0.28  & 0.36  & -0.08             & 0.32              & 1.94          & 9.94              & 13.52             & 0.36  & 0.43  & -0.07             & 0.40              & 17.08             & 48.44             & 57.10             \\ \hline
    (5) SAMP          & 0.22  & 0.30  & -0.08             & 0.26              & \textbf{4.29} & \textbf{18.46}    & \textbf{23.48}    & 0.30  & 0.45  & -0.15             & 0.38              & \textit{20.36}    & \textbf{60.86}    & \textbf{69.13}    \\
    (6) WAVG          & 0.31  & 0.31  & \textbf{-0.01}    & \textit{0.31}     & 2.76          & 10.74             & 14.10             & 0.41  & 0.42  &\textbf{ -0.01}    & \textit{0.42}     & 16.79             & 48.26             & 57.56             \\
    (7) $\tau_{10}$   & 0.21  & 0.37  & -0.16             & 0.29              & 2.57          & 14.32             & 18.69             & 0.28  & 0.48  & -0.20             & 0.38              & 14.71             & 53.94             & 63.01             \\ 
    (8) $\tau_{100}$  & 0.23  & 0.37  & -0.14             & 0.30              & 2.35          & 12.32             & 16.28             & 0.29  & 0.47  & -0.19             & 0.38              & 14.85             & 52.95             & 62.25             \\
    (9) $\tau_{1000}$ & 0.30  & 0.32  & \textbf{-0.01}    & \textit{0.31}     & 2.67          & 10.94             & 14.20             & 0.38  & 0.42  & -0.04             & 0.40              & 15.33             & 47.87             & 57.38             \\

    \hline 

 \end{tabular}

\caption{The evaluation of different embeddings and both data sets. Within each data set, the first group is trained with GloVe on different subsets,
the second group are embeddings created without and the last group with consideration of the data distribution.  For $\mathcal{J}^{\Delta}$ we are hoping for a value close to $0$, for the analogy test higher values mean better performance.
    The $\mathcal{J}$ measures are described in detail in Section~\ref{sec:evaluation}.  
    Best values are in \textbf{bold}, best values within groups are in \textit{italic}.
}
     \label{tablebig}
   \vspace{-.5cm}
\end{table*}

Considering how embeddings encode word contexts, we illustrate the influence of data subsets on the final embedding on two real world data sets.

\textit{New York Times 1990-2016:}
The New York Times dataset\footnote{\url{https://sites.google.com/site/zijunyaorutgers/}} (NYT) contains headlines and lead texts of news articles published online and offline in the New York Times between 1990 and 2016 with a total of 99.872 documents. Political offices as well as sports teams are very closely discussed based on their representatives players, hot topics and their current score. Their context changes over time. As word embeddings are mainly based on the context of a word, their connotation and vectorial representation are influenced by those changes. We investigate the influence of these changes on common word embeddings by splitting this data set in subsets, the first one reaching from 1990-1999 (33.383 articles) and a second one from 2000-2016 (62.058 articles) 

\textit{English Wikipedia:}
The Wikipedia data set (Wiki) contains articles from the English Wikipedia snapshot from April 1st, 2019. We select 12.236 articles from the category \textit{Arts} as well as 24.473 articles from the category \textit{Politics} to analyse the individual influence of those 2 fields on joint word embeddings.\\

\begin{figure}[h]
        \centering
        \includegraphics[width=\columnwidth]{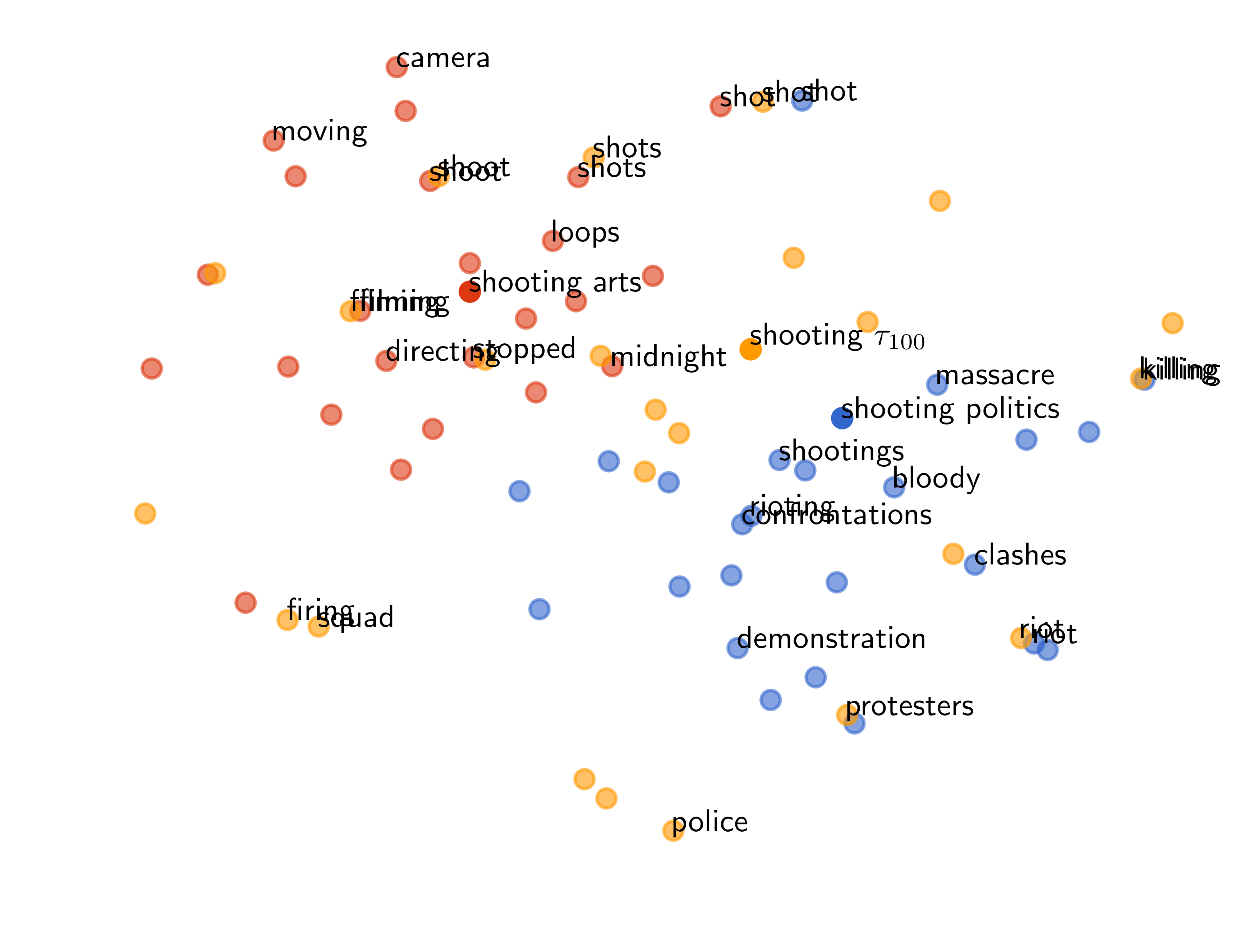}
    \caption{2D tSNE embeddings of the word \emph{shooting} with its NN in different embeddings trained on Wiki: (a), (b), (8) in red, blue and orange respectively.}
    \label{fig:tsne}
\end{figure}

As a first example, we consider the word \textit{shooting} whose nearest neighbors (NNs) in both category groups of the Wiki data set are shown in Fig.\ref{fig:tsne}. Clearly, within Politics, \textit{shooting} refers mostly to the firing of a gun, for Arts, \emph{shooting} rather relates to a photo or movie shooting. When we train embeddings on the joint data set, the new vector reflects both realities, but is biased towards Politics due to the increased number of articles (23/100 and 51/100 common neighbors with the embedding from Arts and Politics, respectively).

Given this intuition, we would like to quantify the retained influence of the data subset (a) and (b) on embeddings $U$.
Inspired by the Jaccard index we compare the neighboring words of a given embedding trained on a subset and those of the composed embeddings $U$.
In more detail, given the sets of $n$ NNs, $\mathcal{N}_n(u^i)$ and $\mathcal{N}_n(v^i)$ for two embeddings $u^i$ and $v^i$ of a word $i$,
the ratio of shared nearest words is:
\begin{equation}
    \mathcal{J}_n(u^i, v^i) = \frac{|\mathcal{N}_n(u^i) \cap \mathcal{N}_n(v^i)|}{n}
\end{equation}
and we denote the average over all words as $\mathcal{J}_n(U, V)$.
For instance $\mathcal{J}_n(U, V) = 0.6$ would mean that words in $U$ and $V$ share on average $60\%$ of their $n$ NNs.
We will use $\mathcal{J}^{s}(U) = \mathcal{J}_n(U_{s}, U)$ and $\mathcal{J}^{l}(U) = \mathcal{J}_n(U_{l}, U)$ to indicate the retained influence of the according subsets on a resulting embedding.
We use the cosine similarity to compute NNs for neighborhoods of different sizes $n$.

\section{Methods}
\label{sec:methods}
We use a number of different embeddings that can be divided into three groups:
merging the data \textit{before} learning the embeddings,
\textit{static} merging algorithms, and \textit{dynamic} merging approaches.

\textit{{Baselines - (a), (b), (1)}}
As baselines we train word embeddings with GloVe \cite{pennington2014} on NYT on articles from (a) 1990-1999 and (b) 2000-2016. 
The resulting embeddings learned on (a), (b), and (1) are denoted as $U_{90}$, $U_{00}$, and $U_{90/00}$.
We further trained word embeddings with GloVe on Wiki for

(a) \textit{Arts} ($U_{arts}$), (b) \textit{Politics} ($U_{pol}$) and (1) the merged data ($U_{a/p}$).\\ GloVe embeddings are trained as 50-dimensional word embeddings on both NYT and Wiki with $x_{\max}=100$, $\alpha=0.75$. We choose a context window size of 15 for NYT and 5 for Wiki as the data set is considerably larger than NYT. We select one vocabulary for each data set and consider only words that occur at least 40 (NYT) and 250 (Wiki) times in the whole data set which leads to vocabularies of size 21398 (NYT) and 19936 (Wiki).

\textit{{Static merging - (2), (3), (4)}}
In constrast to (1) --- merging before learning --- the following approaches merge trained embeddings.
They were proposed by \citet{goikoetxea2016single}.
Given the embeddings $U_{s}$ and $U_{l}$ of the subsets,
method (2) is to average them, i.e. $0.5 \cdot (U_{s}+U_{l})$,
(3) is to concatenate them to a 100-dimensional embedding,
and (4) extends (3) by extracting the 50 most informative dimensions using PCA.
(3) and (4) obtained good results in \citet{goikoetxea2016single}.

\textit{{Dynamic merging - (5), (6), (7), (8), (9)}}
We found that previously presented embeddings are biased towards the larger subset: $\mathcal{J}_{s} << \mathcal{J}_{l}$.
To alleviate this we propose the following approaches.
A first attempt (5) is to upsample the smaller subset to the same size of the larger set. This leads to embeddings with a high score in analogy tests but a decrease in $\bar{\mathcal{J}}$.
We further intent to balance the impact of the subsets by taking an average that is weighted by their inverse proportions (6): $U_{wavg} = 0.65 \cdot U_{s}+ 0.35 \cdot U_{l}$.\\ 
Unfortunately, we found that this approach results in embeddings with inferior quality.
We define an optimization problem that on one hand optimizes the GloVe loss to obtain qualitative good embeddings and
on the other hand tries to balance the influence of the respective subsets by regularizing the distance of the solution to the weighted embeddings $U_{wavg}$.
Given the co-occurence matrix $Y$ and the GloVe weighting function $f(Y)$~\cite{pennington2014} the embeddings $U$ are created by optimizing: 
\vspace{-.2cm}
\begin{align*}
\label{thequation}
\argmin_A  ~~ & ~~ f(Y) \odot \norm{U U^\top - \log Y}^2 \\
    &~~+~\tau ~ \norm{U - U_{wavg}}^2  \tag{2} \\[2ex]
\text{where}~~ U &= A \odot U_{s/l} + (1-A) \odot U_{wavg} \\
A_{i, j} &\in [a_{min}, a_{max}], 0 \leq a_{min} < a_{max} \leq 1
\end{align*}
and $\odot$ denotes a point-wise multiplication.
The regularization parameter $\tau$ allows to trade-off between embedding quality and a balanced influence.
We restrict the solution space to the "rectangle" between $U_{s/l}$ and $U_{wavg}$ and leave exploring an unconstrained version to future work. We optimize Eq.\ \ref{thequation} with gradient descent. We therefore use Adam with a learning rate of $1\mathrm{e}{-3}$ and default values for $\beta$. The optimization is stopped after $10 000$ steps. We have implemented this in PyTorch.
\begin{table}
\begin{adjustbox}{max width=\columnwidth}
  \centering
  \begin{tabular}{| c | c | c || c | c | c  | }
    \hline
    (a) 90 & 90/00 & W-AVG & (b) 00 & 90/00 & W-AVG\\
    \hline
    \hline
    \textbf{war} & 0 & 0 & \textbf{war} & 0 & 0\\
    \hline
    \textbf{vietnam} & 3 & 1 & \textbf{ii} & 2 & 2 \\
    \hline
    persian & 5 & 3 & irag & 1 & 6 \\
    \hline
    gulf & 9 & 4 & \textbf{vietnam} & 3 & 1 \\
    \hline
    era & 17 & 5 & fight & 4 & 10 \\
    \hline
    bidding & - & 7 & combat & 13 & 11 \\
    \hline
    \textbf{ii} & 2 & 2 & wag & - & 21 \\
    \hline
    veteran & - & 8 & terrorism & 18 & 15 \\
    \hline
    cold & 16 & 9 & enemy & 21 & 22 \\
    \hline
    confrontaion & - & - & hero & - & - \\
    \hline
    capture & - & 16 & invasion & 12 & - \\
    \hline 
    \end{tabular}

\end{adjustbox}
    \caption{10 NNs of the word \textit{war} are displayed for $U_{90}$ and $U_{00}$ in column 1 and 4 (NYT). In column 2 and 3 one can see the position the respective word gets after merging for $U_{90/00}$ and W-AVG. On the right side of the table we did the same for $U_{00}$.
    }
    \label{war} 
    \vspace{-.5cm}
\end{table}
\section{Results}

\begin{figure}[t!]
\begin{center}
  \centering
  \includegraphics[width=0.83\linewidth]{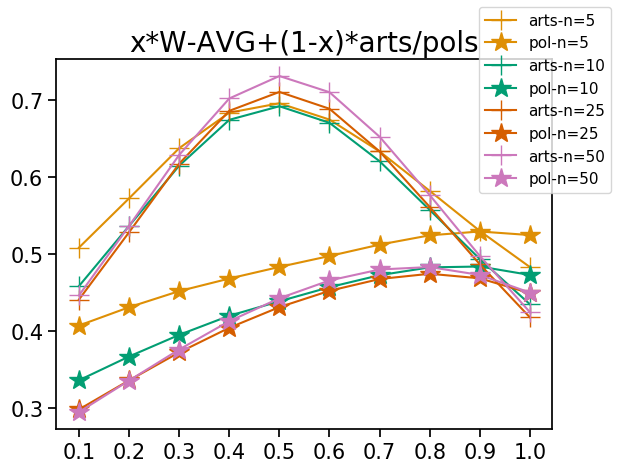}
  \caption{
 Values for $\mathcal{J}_n^{arts}$ and $\mathcal{J}_n^{pol}$  for $n \in \{5, 10, 25, 50\}$ for different weighting parameters of \\$x \cdot$ W-AVG$  + (1-x) \cdot U_{a/p}$ }
  \label{figurebig}
  \end{center}
  \vspace{-.6cm}
\end{figure}

We evaluate the quality of the obtained embeddings $U$ by measuring their performance on analogy tests \cite{mikolov2013b}
and how the influence of the subset is balanced by measuring the number of common neighbors $\mathcal{J}_n^{s}(U)$, $\mathcal{J}_n^{l}(U)$, their average $\bar{\mathcal{J}}$ and their difference $\mathcal{J}^{\Delta} = \mathcal{J}^{s}(U)-\mathcal{J}^{l}(U)$ (see Section~\ref{sec:evaluation}).
$J^{s}$ and $J^{l}$ indicate the respective average over $n \in \{5, 10, 25, 50\}$ for slices $s$(mall) and $l$(arge). Results for all methods, evaluated and averaged over the entire vocabulary, are summarized in Table~\ref{tablebig}.

\textit{Balanced influence:}
First we note that embeddings of the subsets (a) $U_{s}$ and (b) $U_{l}$ have only few NNs in common.
Furthermore, when trained on both subsets the embeddings (1) show a clear shift towards the larger subset (b).
Qualitatively this can also be observed in Table~\ref{war} where we depict the $10$ NNs for the word "war" in subset (a) and subset (b); and the position of the word in the ranked neighbors of (1) in the column "90/00".
We observe that most of the NNs of (a) are not present in the first $10$ NNs of (1),
while for (b) the set of 4 NNs is identical with (1).
Moreover, we note that also the static merging approaches (2), (3) \& (4) exhibit the same shift (see Table~\ref{tablebig}).\\
We try to increase the influence of $U_s$ by upsampling (5) the data set to the size of $U_l$ before training GloVe embeddings. This leads to the same (or even better) quality of the word embeddings as (1) but also results in a decreased $\bar{\mathcal{J}}$.
To alleviate this we propose a weighted average (6) in order to consider the subset proportions.
The results in Table~\ref{tablebig} indicate that this simple approach indeed yields, in terms of our measure, balanced embeddings.
This can also be observed exemplary in Table~\ref{war} where NNs of (6) correlate much more with the NNs of the respective subsets.

Unfortunately, we will see that the embedding quality suffers when performing a weighted average.
With the aim to align both desiderata --- balanced influence of the subsets and quality of the embeddings --- we proposed an optimization procedure (7-9).
From Table~\ref{tablebig} we read that the resulting embeddings for different regularization strengths $\tau$ are balanced, but surprisingly the influence of the respective subsets decreases in comparison to (1).
As a control experiment we consider the embeddings given by a weighted average between (1) and (6) (Figure~\ref{figurebig}),
where this drop of influence cannot be noted.
Yet none of the such averaged embeddings yields good performance and balancing; which justifies the application of an optimization procedure.

\textit{Embedding quality:}
We measure the embedding quality by means of analogy tests.
The embeddings trained on all the data (1) perform best in this context --- hinting that it is beneficial to leverage as much information from data as possible.
The statically merged embeddings (2), (3), (4) do not perform as well on our task, in contrast to the results of \citet{goikoetxea2016single}.\\
Furthermore, we note that the weighted average (6) results also in a decrease in embedding quality.
In contrast, we find that our optimization approach is able to capture both, embedding quality and balances the influence of the subsets.

\section{Discussion}
Considering that text corpora are often composed of subsets,
embedding learners merge them in \textit{incidental} manner
--- either by merging the text before or the word vectors after
training.
We argue this can lead to undesired shifts in the embedded semantics
and propose a measure for this shift as well as approaches to balance the composition of the subsets.

Our preliminary results show that one can indeed level the impact of different subsets.
A weighted average of the subset embeddings yields balanced word embeddings, yet their quality decreases.
The proposed optimization routine results in word vectors with good quality and balanced, yet decreased influence of the subsets.

As future work we aim to extend our empirical results and investigate the proposed optimization routine in more detail, e.g., by removing the constraints.
As additional experiments we would like to investigate the influence of the different combination methods on downstream tasks, such as classification of sub-categories of the Wikipedia articles. This will further our understanding of the workings of the combination methods in comparison to the analogy tests that are not data slice specific.
As alternative to the current regularization --- that minimizes the distance to another, presumably balanced embedding --- we would like to develop a (differentiable) regularization term that is closer related to our measure $\mathcal{J}_n^{s}(U)-\mathcal{J}_n^{l}(U)$.
Adapting the work of \citet{Berman2018TheLL}, which proposes surrogate losses for the Jaccard index, seems to be a promising direction for this goal.

An interesting question posed by our results is how merging of data subsets impacts the resulting embedding semantics
--- considering that many NNs of $U_{s/l}$ are not NNs for the subset embeddings $U_{s}$ and $U_{l}$.

\section*{Acknowledgments}
This work was supported by the Federal Ministry of Education and Research (BMBF) 
for the Berlin Big Data Center BBDC (01IS14013A) and for the MALT III project (01IS17058). We thank L. Ruff, T. Schnake, O. Eberle and S. Dogadov  for fruitful discussions. We also thank the reviewers from ACL and the Workshop on Ethical, Social and Governance Issues in AI at NeurIPS 2018 for their valuable comments.

\bibliography{main}

\begin{thebibliography}{10}
\expandafter\ifx\csname natexlab\endcsname\relax\def\natexlab#1{#1}\fi

\bibitem[{Berman et~al.(2018)Berman, Triki, and Blaschko}]{Berman2018TheLL}
Maxim Berman, Amal~Rannen Triki, and Matthew~B. Blaschko. 2018.
\newblock The lovasz-softmax loss: A tractable surrogate for the optimization
  of the intersection-over-union measure in neural networks.
\newblock \emph{2018 IEEE/CVF Conference on Computer Vision and Pattern
  Recognition}, pages 4413--4421.

\bibitem[{Goikoetxea et~al.(2016)Goikoetxea, Agirre, and
  Soroa}]{goikoetxea2016single}
Josu Goikoetxea, Eneko Agirre, and Aitor Soroa. 2016.
\newblock Single or multiple? combining word representations independently
  learned from text and wordnet.
\newblock In \emph{Thirtieth AAAI Conference on Artificial Intelligence}.

\bibitem[{Henriksson et~al.(2014)Henriksson, Moen, Skeppstedt,
  Daudaravi{\v{c}}ius, and Duneld}]{Henriksson2014}
Aron Henriksson, Hans Moen, Maria Skeppstedt, Vidas Daudaravi{\v{c}}ius, and
  Martin Duneld. 2014.
\newblock \href {https://doi.org/10.1186/2041-1480-5-6} {Synonym extraction and
  abbreviation expansion with ensembles of semantic spaces}.
\newblock \emph{Journal of Biomedical Semantics}, 5(1):6.

\bibitem[{Joulin et~al.(2016)Joulin, Grave, Bojanowski, and
  Mikolov}]{joulin2016}
Armand Joulin, Edouard Grave, Piotr Bojanowski, and Tomas Mikolov. 2016.
\newblock Bag of tricks for efficient text classification.
\newblock \emph{arXiv preprint arXiv:1607.01759}.

\bibitem[{Mikolov et~al.(2013{\natexlab{a}})Mikolov, Chen, Corrado, and
  Dean}]{word2vec}
Tomas Mikolov, Kai Chen, Greg Corrado, and Jeffrey Dean. 2013{\natexlab{a}}.
\newblock \href {http://arxiv.org/abs/1301.3781} {Efficient estimation of word
  representations in vector space}.
\newblock \emph{CoRR}, abs/1301.3781.

\bibitem[{Mikolov et~al.(2013{\natexlab{b}})Mikolov, Le, and
  Sutskever}]{mikolov2013b}
Tomas Mikolov, Quoc~V. Le, and Ilya Sutskever. 2013{\natexlab{b}}.
\newblock Exploiting similarities among languages for machine translation.
\newblock \emph{CoRR}, abs/1309.4168.

\bibitem[{Miller(1995)}]{Miller1995}
George~A. Miller. 1995.
\newblock \href {https://doi.org/10.1145/219717.219748} {Wordnet: A lexical
  database for english}.
\newblock \emph{Commun. ACM}, 38(11):39--41.

\bibitem[{Pennington et~al.(2014)Pennington, Socher, and
  Manning}]{pennington2014}
Jeffrey Pennington, Richard Socher, and Christopher~D. Manning. 2014.
\newblock \href {http://www.aclweb.org/anthology/D14-1162} {Glove: Global
  vectors for word representation}.
\newblock In \emph{Empirical Methods in Natural Language Processing (EMNLP)},
  pages 1532--1543.

\bibitem[{Rothe and Sch{\"u}tze(2017)}]{rothe2017autoextend}
Sascha Rothe and Hinrich Sch{\"u}tze. 2017.
\newblock Autoextend: Combining word embeddings with semantic resources.
\newblock \emph{Computational Linguistics}, 43(3):593--617.

\bibitem[{Speer and Chin(2016)}]{speer2016ensemble}
Robert Speer and Joshua Chin. 2016.
\newblock An ensemble method to produce high-quality word embeddings.
\newblock \emph{arXiv preprint arXiv:1604.01692}.

\end{thebibliography}
\bibliographystyle{acl_natbib}

\end{document}